\documentclass[letterpaper]{article} 
\usepackage{aaai25}  
\usepackage{times}  
\usepackage{helvet}  
\usepackage{courier}  
\usepackage[hyphens]{url}  
\usepackage{graphicx} 
\usepackage{natbib}  
\usepackage{caption} 
\usepackage{algorithm}
\usepackage{algorithmic}
\usepackage{caption}
\usepackage{makecell,amsmath, amssymb, cleveref,algorithm,color,colortbl}
\usepackage{graphicx,multirow,xspace,subcaption,longtable}
\usepackage{booktabs}
\usepackage{adjustbox}
\usepackage{subcaption}

\definecolor{LightRed}{rgb}{1, 0.8, 0.8}
\definecolor{LightGreen}{rgb}{0.8, 1, 0.8}

\usepackage{xcolor}
\usepackage{newfloat}
\usepackage{listings}
\DeclareCaptionStyle{ruled}{labelfont=normalfont,labelsep=colon,strut=off} 
\floatstyle{ruled}
\newfloat{listing}{tb}{lst}{}
\floatname{listing}{Listing}

\newcommand{\eg}{e.g.,\ }
\newcommand{\etal}{et al.\xspace}
\newcommand{\ie}{i.e.,\ }

\newcommand{\g}[1]{$\mathcal{G}_{#1}$}
\newcommand{\ext}{$\mathcal{G}_{ext}$\ }
\newcommand{\neu}{$\mathcal{G}_{neu}$\ }

\newcommand{\cext}{$\mathcal{C}_{high}$\ }
\newcommand{\cneu}{$\mathcal{C}_{low}$\ }

\newcommand{\para}[1]{{\bf \noindent #1 }}
\newcommand{\CP}{C$\rightarrow$ P\ }
\newcommand{\PC}{P$\rightarrow$ C\ }

\newcommand{\answerYes}[1]{\textcolor{blue}{#1}} 
\newcommand{\answerNo}[1]{\textcolor{teal}{#1}}
\newcommand{\answerNA}[1]{\textcolor{gray}{#1}} 

\newcommand{\update}[1]{\textcolor{blue}{#1}}
\renewcommand{\update}[1]{#1}

\title{Examining the Role of YouTube Production and Consumption Dynamics on the Formation of Extreme Ideologies}
\author{
    Sarmad Chandio,
    Rishab Nithyanand
}
\affiliations{
    University of Iowa \\
    \{sarmad-chandio, rishab-nithyanand\}@uiowa.edu 
}

\begin{document}

\maketitle

\begin{abstract}
The relationship between content production and consumption on algorithm-driven platforms like YouTube plays a critical role in shaping ideological behaviors. While prior work has largely focused on user behavior and algorithmic recommendations, the interplay between what is produced and what gets consumed, and its role in ideological shifts remains understudied. In this paper, we present a longitudinal, mixed-methods analysis combining one year of YouTube watch history with two waves of ideological surveys from 1,100 U.S. participants. We identify users who exhibited significant shifts toward more extreme ideologies and compare their content consumption and the production patterns of YouTube channels they engaged with to ideologically stable users. Our findings show that users who became more extreme consumed have different consumption habits from those who do not. This gets amplified by the fact that channels favored by users with extreme ideologies also have a higher affinity to produce content with a higher anger, grievance and other such markers. Lastly, using time series analysis, we examine whether content producers are the primary drivers of consumption behavior or merely responding to user demand.

\end{abstract}

\section{Introduction} \label{sec:Intro}

\para{Platform algorithms have reshaped the relationship between content creators and audiences.}
On modern platforms, algorithms are not just responsible for curating content for their users. They are also responsible for operationalizing the incentives that govern what content gets produced \cite{sandvig2014auditing}.
Consequently, the mechanisms through which information is produced and consumed have undergone a drastic shift over the past decade.
For example, content creators now rely on favorable perceptions from the platform's recommendation and monetization algorithms \cite{dunna2022paying} and thus tinker with their content to improve its reach, visibility, and engagement. 
In this system where engagement is paramount, content consumers have powers not seen in the traditional media ecosystem: their direct engagement (or lack thereof) directly influences the ability of a creator to reach other consumers. 
As a result, audience engagement becomes a feedback signal not just for what the audience enjoy but also for what producers must create. 
This is an inversion the producer-consumer dynamics that are apparent in the supply-driven traditional media ecosystems where producers created and controlled their own narratives without fear of the algorithmic imaginary \cite{schulz2023new} and audiences selected from a small number of consumption options.

\para{Understanding the modern content producer-consumer influence relationship is critical for effective platform  governance and understanding how extremist ideologies propagate.}
In the context of polarizing or problematic content, prior empirical research has mostly focused on the behaviors of and relationships between algorithms and content consumers. For example, they have identified the characteristics of content that algorithms favor \cite{haroon2022youtube, chandio2024audit} and established that users within these ecosystems exhibit tendencies towards selective exposure \cite{habib2024uncovering}.
However, fully understanding the modern social media landscape requires us to investigate a critical, yet empirically understudied, question related to the relationship between content producers and consumers: {\em Do content producers drive extreme ideologies and narratives to their audiences, or does their need to maximize engagement and algorithmic favor lead to them reflecting the ideologies of their audiences?} 
Answering this question requires us to assess production, consumption, and behavioral patterns that can clarify the nature and direction of influence.
This is challenging because we currently lack high-fidelity longitudinal data that can discern whether the spread of extreme rhetoric and ideologies occurs because of the supply-side (creators leading the way) or demand-side (creators reflecting what their audiences wish to consume).
However, without understanding who drives extreme ideologies we risk creating ineffective governance and intervention strategies which ignore behavioral dynamics within the media ecosystem. 

\para{This empirical study is a step towards understanding the producer-consumer influence relationship on social media.} 
Specifically, this paper examines whether the rise of algorithm-driven platforms has inverted the traditional producer-consumer relationship.
To do so, we analyze data gathered from a mixed-methods approach. This data from 1.1K participants within the US include a two-wave survey, each wave one year apart, of their attitudes towards five hot-button sociopolitical issues (race, vaccines, immigration, and abortion) and their YouTube watch histories over that duration. We leverage this data to investigate three research questions.

\begin{itemize}
    \item {\em RQ1. What are the measurably different markers of content consumed by individuals who became more ideologically extreme compared to those who did not? (\Cref{sec:rq1})} We use the survey to identify individuals who demonstrated a significant shift towards more extreme opinions on at least one issue and assess if and how their content consumption habits differ from those that did not. We find that users who became more ideologically extreme over the period of our study consumed content that generally exhibited substantially higher levels of anger and grievance.

    \item {\em RQ2. What distinguishes content that is associated with increasing extremism? (\Cref{sec:rq2}}) Here, we shift our focus to analyzing the characteristics of the content produced by YouTube channels that were disproportionately popular among each of our user groups (\ie those that did and did not experience a significant shift towards more extreme ideologies). We analyze if and how the markers varied within similar types of content (\eg news) that had the largest affinity with each group. We find that channels with high affinity to users who became more extreme consistently produced content with significantly higher levels of anger, power, negativity, and grievance. The most pronounced difference was in anger (+39.8\%), suggesting that these users were embedded in a distinct media ecosystem.

    \item {\em RQ3. What is the direction of the influence relationship between the content that is being produced and the content that is being consumed? (\Cref{sec:rq3}}) Here, we model the temporal dynamics between content supply (produced content) and demand (broader audience consumption behaviors) to determine whether the ecosystem is supply-driven (producers influencing users) or demand-driven (users influencing producers). We find that content production remains the dominant force shaping consumption behavior. Further, anger exhibits a bidirectional relationship in the extreme group, indicating that consumers are also influencing production choices.
\end{itemize}

\section{Data and methods} \label{sec:data}
At a high-level, our goal is to understand whether extreme ideological shifts among users can be explained by the content that they chose to consume or the content that was produced for them.
To this end, we develop datasets and methods suitable for characterizing users as having undergone an extreme ideological shift and for extracting features from videos (that they consumed or were produced for them).
In this section, we describe the data and methods.
We begin by outlining our two-wave survey used to classify users as having extreme or neutral ideological shifts (\Cref{sec:data:survey}). 
Next, we describe the characteristics of the consumption history behavioral data from users in our extreme and neutral groups and detail our approach for extracting features from YouTube videos (\Cref{sec:data:features}).
We provide details about how these features were used to answer each of our research questions in their respective sections.

\subsection{Survey-based categorization of participants}\label{sec:data:survey}
\para{Participant demographics and survey overview.}
We fielded a two-wave survey of 1,100 US adults recruited via YouGov's Pulse panel \cite{2024AboutYouGov}. Participants were screened for minimum engagement with current events (they reported following the news at least ``sometimes''). The demographic characteristics (age, race, education, religiousness, and income) and political affinity of this population reflected the demographics of the US voting population.
The first wave of the survey was fielded in August 2021 and the second wave in August 2022. Each wave measured beliefs and attitudes across five sensitive sociopolitical issues: race, abortion, immigration, vaccines, and climate change.

\begin{figure}[th]
    \centering
    \includegraphics[width=.48\textwidth, trim={.2cm 0cm .2cm 1.1cm}]{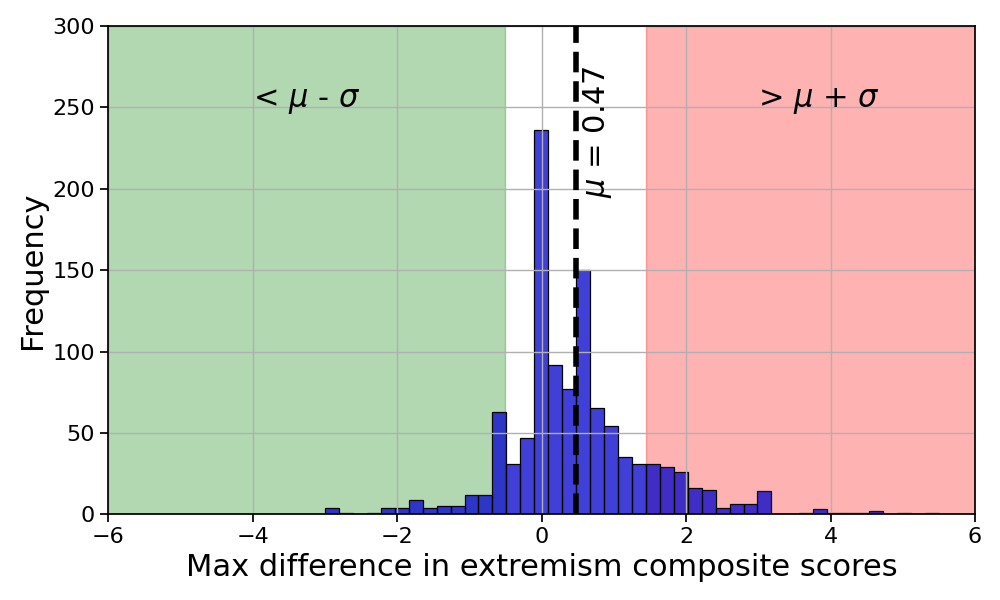}
    \caption{Distribution of maximum shifts in ideological attitudes and beliefs between the two waves.}
\label{fig:distribution}
\end{figure}

\para{Identifying participants with shifts towards extremism.}
To quantify ideological extremity, we adapted a behavioral radicalism scale based on prior work by Moskalenko and McCauley (\citeyear{mccauley2008mechanisms}). For each issue, participants rated their agreement (on a seven-point Likert scale) with four statements expressing willingness to support violent or illegal actions aligned with their own views and beliefs. 
For example, the abortion scale included the items: {\em ``I would support an organization that fights for my beliefs about abortion even if it sometimes breaks the law''} and {\em ``I would continue to support an organization that fights for my beliefs about abortion even if they sometimes resort to violence''}. 
For each wave and for each issue, these items were used to create a composite (average) score associated with a participants tendency towards extremism.
Each participant in our survey was randomly assigned two of the five issues and these assigned issues were consistent for both waves.
To identify whether a participant developed more extreme ideologies about the issues they were assigned, we calculated the difference in their composite scores between the two waves for each issue, 
\update{ and then selected the maximum change across those issues for use in our analysis\footnote{\update{While we observe shifts across multiple issues, our sample size is not sufficient to statistically analyze content-issue alignment at the per-issue level.}}. This ensures that participants are classified based on the most substantial ideological shift they experienced, rather than averaging or diluting shifts across multiple issues (read \Cref{appendix:specific-issues} for issue-specific shifts in attitudes).}
On average, our participants demonstrated a marginal shift towards extremism between the two surveys. The mean ($\mu$) and standard deviation ($\sigma$) of this shift were 0.47 and 0.98, respectively. \Cref{fig:distribution} shows how these shifts were distributed.
Participants whose highest composite score difference was more than one standard deviation above the population mean (\ie $> \mu + \sigma$) taken across all issues were classified as having undergone a substantial ideological shift, and labeled as {\em more extreme}. 158 (14.4\%) of our participants were assigned this label.
Those who had a shift that was within one standard deviation of the population mean (\ie within [$\mu - \sigma$, $\mu + \sigma$]) were classified as {\em unchanged}. 855 (77.8\%) of our participants were assigned this label.

\subsection{YouTube consumption and feature extraction} \label{sec:data:features}

\para{YouTube video consumption data.} In parallel with our survey, we obtained one year of YouTube video consumption history for all participants over the duration between the two survey waves. 
This was obtained via YouGov's passive tracking infrastructure. Participants could temporarily disable tracking at will; since these events are not marked, we conservatively excluded users who were found to have watched fewer than 50 videos, after filtering for ads.
In total, this left us with 52 participants with the {\em more extreme} label and 286 participants with the {\em unchanged} label. 
These users and their consumption histories serve the basis for the analysis and results presented in the remainder of this paper. We refer to the 52 {\em more extreme} participants as \ext and the 286 {\em unchanged} participants as \neu.
In total, these participants consumed 617K videos from 56.6K unique channels with participants in \ext consuming 75K videos ($\mu$: 1447 videos/participant) from 14.9K unique channels and those in \neu consuming 541K videos ($\mu$: 1894 videos/participant) from 46.9K unique channels. Although both groups had a small number of users who exhibited disproportionately high viewing volumes, we ensure in our analysis that each user is weighted equally to ensure that these outliers do not skew our findings.
An examination of the categories of these videos showed that three categories ({\em People \& Blogs}, {\em News \& Politics}, and {\em Entertainment}) accounted for nearly half of the total viewing for each group. We also observed that both groups had different watch interests. For example, \ext participants were found to consume more content in the {\em Gaming}, {\em Science \& Technology}, and {\em Vehicles} categories, while \neu participants were found to consume more content in the {\em Comedy}, {\em Music}, and {\em Movies} categories.

\para{Video feature extraction.} To analyze content-level trends that might be associated with extremism-related attitudes, we extracted linguistic markers from the videos consumed by participants.
For each video analyzed in this project, we used the YouTube API to download the title, channel, and transcript. After standard pre-processing (lower-casing, removing stop words, and lemmatization), we applied the extremism-related feature extractors adapted from Habib \etal (\citeyear{habib2022making}), representing the the language of the video (title + transcript) as a 6-dimensional feature vector.
These methods draw from behavioral threat assessment frameworks in psychology and criminology \cite{trap-18, elaine2012calibrating, erg-22, cole2010guidance} to detect language associated with `{\em warning behaviors}' of extremism.
These warning behaviors include grievance (the feeling of injustice, unfair treatment, and frustration over suffering), power (the need for having impact, control, and influence over others), anger, negative outlook or sentiment, and in-/out-group identification\footnote{\update{Anger, fixation, in-group/out-group, and power are based on LIWC dictionary \cite{pennebaker2015development}. Grievance is measured using the Grievance Dictionary \cite{van2021grievance}. Sentiment is computed using the VADER sentiment analyzer \cite{hutto2014vader}.}
}.
Using their approaches for measuring these traits, we created a feature vector ($F_i$ = $\langle f_{i1}, \dots, x_{i6}\rangle$) for each video ($v_i$). Here, each feature $f_{ij}$ represents a measure of the prevalence of language within the video $i$ that can be associated with the trait $j$.

\subsection{Ethical considerations} \label{sub:ethics}
This study and its associated data collection procedures were approved by our Institution Review Board. All participants consented to the terms of this research. Consent and compensation were managed by YouGov, our data collection partner for this project. 
Participants were allowed to turn off monitoring their online activities through the Pulse extension and app. 
We only received anonymized behavioral and survey data from YouGov, with no identifiers that could be used to link the data back to their creators. To mitigate the privacy risks of study participants, we do not plan to make our data publicly available.
Metadata of YouTube videos gathered as part of this study were obtained using the YouTube API while respecting their API rate limits.

\section{Characterizing consumption behaviors} \label{sec:rq1}

In this section, we focus on understanding whether individuals who developed more extreme ideological views over the course of the study consumed content with different linguistic characteristics compared to individuals whose views remained stable.
Specifically, {\em RQ1. What are the measurably different markers of content consumed by individuals who became more ideologically extreme compared to those who did not?}
This question is motivated by the need to empirically evaluate whether these individuals (in \ext) were immersed in informational environments that differed from others (in \neu).
We first describe the methods related to this specific investigation (\Cref{sec:rq1:methods}) and then describe our findings (\Cref{sec:rq1:results}).

\subsection{Methods} \label{sec:rq1:methods}
We approach this research question in two parts. First, we compare aggregate consumption behaviors across the two groups (\neu and \ext). Second, we examine how each group's consumption behaviors changed over the course of our study. 

\para{Comparing consumption behaviors.}
We performed three comparisons of features in consumed content: 
(1) \neu with \ext over the entire year; 
(2) \neu during each pair of sequential quarters (\eg \neu Q1 vs. \neu Q2); and
(3) \ext during each pair of sequential quarters.
The first comparison allows us to understand aggregate differences between the two groups over the period of the entire year, while the last two comparisons shed light on how consumption behaviors of each group evolved over time.
%

\para{Creating and comparing distributions of extremism markers.}
For each group \g{k} within a comparison, we create a boolean-valued matrix $M$ such that $M_{ij}$ indicates whether the user $u_i$ in the group consumed video $v_j$.
Then, to compare consumption behavior between \neu and \ext, we use a bootstrapped sampling method that ensures equal representation of users regardless of their overall watch volume. We do this to mitigate the influence of the heavy-volume outliers described earlier.
Specifically, for each group (\neu and \ext), we repeat the following process 10K times:
First, we randomly select a row $i$ from $M$. This row is associated with the watch history of the user $u_i$. Next, we randomly select a $j$ such that $M_{ij}=1$. This effectively selects one video ($v_j$) that was consumed by $u_i$. Finally, we associate the feature vector $F_j$ with the group.
At the end of this process, each group is associated with a collection of 100K feature vectors that reflect the content consumed by the users (and time periods) that they represent, without over-representing the consumption habits of our outliers.
This collection generates an empirical distribution of warning behavior marker values for each group. Put differently, they are the distribution of the prevalence of an extremism-related linguistic trait seen within the content that group members consumed.
Then, to compare the differences in the prevalence of a specific marker between \neu and \ext, we use a two-sample $t$-test and report the difference in their means.



\begin{figure*}[th!]
    \centering
    \includegraphics[width=1\linewidth]{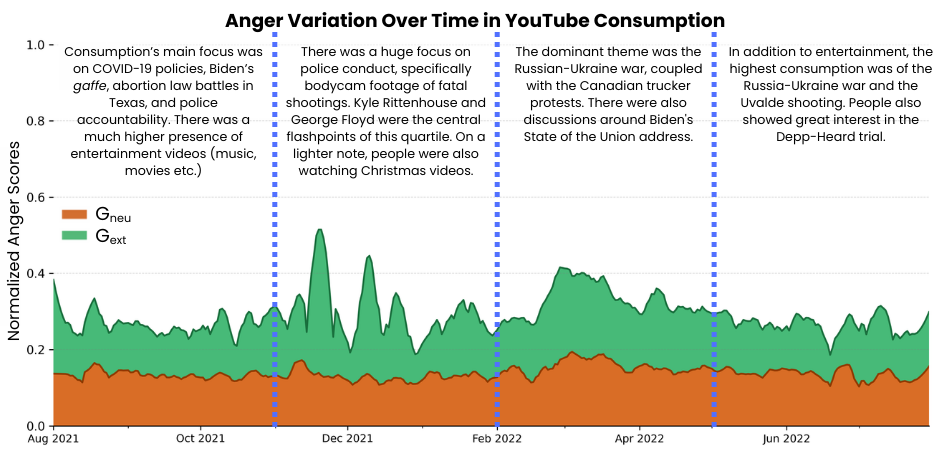}
    \caption{This figure shows the temporal evolution of daily averaged anger scores for \ext and \neu over the year. The blue lines divide the timeline into four distinct quartiles. Read \Cref{appendix:topic-model} in the Appendix for more detail on how the topic summaries for each quartile were computed.}

    \label{fig:topics}
\end{figure*}

\subsection{Results} \label{sec:rq1:results}
To evaluate whether users who developed more extreme views consumed distinct types of content compared to those whose views remained stable, we analyzed both baseline differences and temporal changes in linguistic markers across the two groups. Our results are summarized in \Cref{tab:rq1-users-compact}.

\begin{table}[t]
\centering
\small
\setlength{\tabcolsep}{2pt}
\begin{tabular}{l c  ccc | ccc}
\toprule
\textbf{Marker} & \textbf{$\Delta_{\text{across}}$} &
\multicolumn{3}{c}{$\Delta_{\text{within}}^{\text{neu}}$} &
\multicolumn{3}{c}{$\Delta_{\text{within}}^{\text{ext}}$} \\
\cmidrule(lr){3-5} \cmidrule(lr){6-8}
& & Q2-Q1 & Q3-Q2 & Q4-Q3 & Q2-Q1 & Q3-Q2 & Q4-Q3 \\
\midrule
Anger       & \cellcolor{LightRed}6.8\textsuperscript{*}  & 
\cellcolor{LightRed}0.9\textsuperscript{*} & 
\cellcolor{LightRed}20.5\textsuperscript{*} & 
\cellcolor{LightGreen}–8.5\textsuperscript{*} &
\cellcolor{LightGreen}–17.0\textsuperscript{*} & 
\cellcolor{LightRed}5.9\textsuperscript{*} & 
\cellcolor{LightGreen}–17.3\textsuperscript{*} \\

Griev.      & \cellcolor{LightRed}13.0\textsuperscript{*} & 
\cellcolor{LightGreen}–4.1\textsuperscript{*} & 
0.0 & 
\cellcolor{LightRed}5.4\textsuperscript{*} &
\cellcolor{LightGreen}–12.5\textsuperscript{*} & 
\cellcolor{LightGreen}-17\textsuperscript{*} & 
\cellcolor{LightGreen}–13.1\textsuperscript{*} \\

Power       & \cellcolor{LightRed}0.6\textsuperscript{*}  & 
\cellcolor{LightRed}5.4\textsuperscript{*} & 
\cellcolor{LightRed}12.7\textsuperscript{*} & 
\cellcolor{LightGreen}–14.3\textsuperscript{*} &
\cellcolor{LightGreen}–6.1\textsuperscript{*} & 
\cellcolor{LightRed}2.0\textsuperscript{*} & 
\cellcolor{LightRed}5.5\textsuperscript{*} \\

Neg.        & \cellcolor{LightRed}1.8\textsuperscript{*}  & 
\cellcolor{LightGreen}-2.1\textsuperscript{*} & 
\cellcolor{LightRed}1.3\textsuperscript{*} & 
\cellcolor{LightGreen}-0.2\textsuperscript{*} &
\cellcolor{LightRed}4.5\textsuperscript{*} & 
\cellcolor{LightRed}0.4\textsuperscript{*} & 
\cellcolor{LightRed}0.2\textsuperscript{*} \\

In-group    & 0.0 & 
\cellcolor{LightRed}2.7\textsuperscript{*} & 
\cellcolor{LightGreen}–14.8\textsuperscript{*} & 
\cellcolor{LightGreen}–2.8\textsuperscript{*} &
\cellcolor{LightGreen}–25.3\textsuperscript{*} & 
\cellcolor{LightGreen}–3.7\textsuperscript{*} & 
\cellcolor{LightGreen}–11.3\textsuperscript{*} \\

Out-group   & \cellcolor{LightGreen}–5.8\textsuperscript{*} & 
\cellcolor{LightGreen}–17.1\textsuperscript{*} & 
\cellcolor{LightRed}13.3\textsuperscript{*} & 
\cellcolor{LightRed}37.3\textsuperscript{*} &
\cellcolor{LightGreen}–3.2\textsuperscript{*} & 
\cellcolor{LightRed}22.2\textsuperscript{*} & 
\cellcolor{LightRed}44.6\textsuperscript{*} \\
\bottomrule
\end{tabular}
\caption{Percentage differences in exposure to linguistic markers. $\Delta_{\text{across}}$ shows the difference between \ext and \neu users over the year \update{(\ext - \neu)}. $\Delta_{\text{within}}$ shows the marginal percentage differences within \neu and \ext groups from one quarter to the next. Asterisks denote $p<0.05$.}
\label{tab:rq1-users-compact}
\end{table}

\para{Group-level differences.}
At baseline, our \ext participants exhibited statistically significantly higher exposure to several key markers than their \neu counterparts during the year-long period of the study. Specifically, exposure to content exhibiting linguistic markers of `Anger' and `Grievance' were 6.8\% and 13\% higher, respectively, and their exposure to out-group identification markers was 5.8\% lower.
These statistically significant differences suggest that participants whose views were becoming more extreme were inhibiting more extreme content environments and that there is likely a positive association between these consumption behaviors and the development of their ideologies.
Other markers (power, sentiment, and in-group identification) only had marginal differences (ranging from 0\% to +1.8\%) and thus harder to meaningfully interpret.

\para{Within-group differences.}
For both groups, examining consumption trends and differences over time provided additional insights into the behaviors of \ext and \neu participants.
We observed significant within-group shifts over time in the prevalence of several content markers, with divergent trajectories between the \neu and \ext groups.
For the anger and grievance markers, although \ext had a far higher baseline than \neu, they also showed a declining trend in exposure throughout the year. Specifically, comparing consumption during Q4 with Q1, the prevalence of anger and grievance dropped by 31\% and 20\%, respectively.
In contrast, \neu showed an increasing in these same markers over time (anger rose by 11\% and grievance by 1\%), although they remained consistently below the \ext group throughout the year.
For the sentiment and out-group markers, both groups saw increasing trends throughout the year, although the increases were more drastic for \ext.
Both groups also exhibited similar increases in the power markers with largely parallel trends over time. 

\para{Takeaways.}
All together, our results show that users who became more ideologically extreme over the period of our study consumed content that generally exhibited substantially higher levels of anger and grievance while all other differences were marginal. 
Examining within-group trends, it appears that the events occurring around Q3 of the study period were associated with sharp increases in consumption markers for participants in both groups with higher impact on participants in \neu.
\section{Characterizing production trends}\label{sec:rq2}
We now turn our attention to understanding content production trends.
Specifically, we ask: {\em RQ2. What are the distinguishing markers of content produced by channels that had high affinity with users who developed more extreme ideological views?}
This question is motivated by the need to determine whether the differences observed in user consumption (RQ1) reflect broader differences in the content produced for those users.
Put differently, we ask whether users are actively selecting videos with specific markers or if content with these markers are being produced for and disproportionately reaching different groups.

\subsection{Methods}\label{sec:rq2:methods}
Similar to our analysis of consumption trends, we first compare aggregate production trends for channels with high affinity to each group with each other and then examine how producers within each group evolved over the study period.

\para{Measuring channel affinity.}
To understand production trends that may be associated with increasing markers of extremism, we first needed to identify channels that have a high affinity with the \ext group --- \ie channels that were highly preferred by the viewers from extreme group.
Accordingly, we defined a {\em viewing intensity} measure ($\gamma$), an {\em expected intensity} measure ($\beta$), and a {\em channel affinity} measure ($\alpha$) for each group and for each channel with at least ten unique viewers in our dataset. 
The viewing intensity reflects the average number of views per group member. This relationship reflected in the equation below. We denote a single channel by $c$ and set of all channels by C.
\begin{equation}
\gamma(\text{c}, \mathcal{G}) =
\frac{\text{\# views received by c from } \mathcal{G}}
     {\text{\# members in } \mathcal{G} \text{ who watched c}}
\label{eq:gamma}
\end{equation}

The expected intensity defines the expected number of views per channel of the group, \ie 
\begin{equation}
\beta(\mathcal{G}) =
\frac{\sum_{c \in \text{C}} \left( \text{\# views received by } c \text{ from } \mathcal{G} \right)}
     {\sum_{c \in \text{C}} \left( \text{\# members in } \mathcal{G} \text{ who watched } c \right)}
\label{eq:beta}
\end{equation}

Then, the channel affinity measure reflects the ratio of the observed viewing intensity from the expected viewing intensity 
--- \ie 
\update{
\begin{equation}
\alpha(c, \mathcal{G}) = \frac{\gamma(c, \mathcal{G})}{\beta(\mathcal{G})}
\label{eq:alpha}
\end{equation}
}

\update{This channel affinity measure captures the degree to which a given channel is skewed towards consumers from \g{}. When $\alpha(c, \mathcal{G})$ = 1, it indicates that the observed viewing behavior for channel $c$ among our \g{} participants did not differ from the expected behavior. When $\alpha(c, \mathcal{G})$ $>$ 1, it indicates that members of \g{} had a higher number of views per user than expected. Thus, higher values of $\alpha$ denote a channel's greater affinity towards the group $\mathcal{G}$.
%
We use $\alpha$(c, \ext) $>$ 1.5 as the threshold for identifying channels with high affinity towards the \ext and $\alpha$(c, \ext) $<$ 1 for low affinity channels. We refer to the set of channels that meet these thresholds as \cext and \cneu, respectively. 
In total, \cext contained 34 unique channels and \cneu contained 203 unique channels.}

\para{Identifying production trends.}
For each channel in \cext and \cneu, we used the YouTube API to obtain titles, descriptions, and transcripts of all videos produced during the study period. This yielded 142K and 334K videos produced by channels in \cext and \cneu, respectively.
Next, for each group $\mathcal{C}$, we created a list $L$ such that $L_{i}$ contained the mean of all the markers of videos produced by the channels of $\mathcal{C}$ on the $i^{th}$ day.
Then, to compare production trends between \cext and \cneu, we used a bootstrapped sampling approach (10K repetitions) on the list $L$ belonging to each $\mathcal{C}$.
This bootstrapped sample of productions from \cext and \cneu provides us with an empirical distribution of the prevalence of warning behavior markers observed among content created by channels with disproportionately high or low affinity towards participants in \ext.
To compare the differences in the prevalence of specific markers in content produced by \cext and \cneu, we use a two-sample t-test and report the differences in their means.

\begin{figure*}[th!]
    \centering
    \includegraphics[width=1\linewidth]{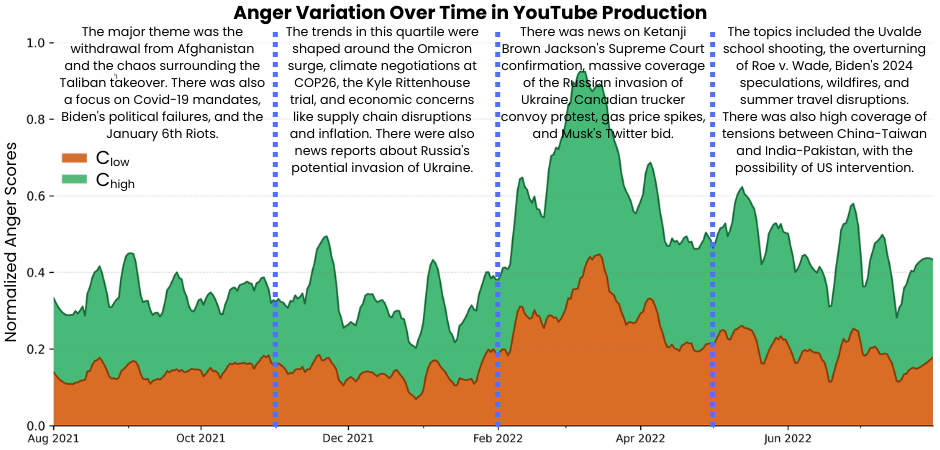}
    \caption{This figure shows the temporal evolution of daily averaged anger scores for \cext and \cneu over the year. The blue lines divide the timeline into four distinct quartiles. Read \Cref{appendix:topic-model} in the Appendix for more detail on how the topic summaries for each quartile were computed.}

    \label{fig:users-anger}
\end{figure*}

\begin{table}[t]
\centering
\small
\setlength{\tabcolsep}{2pt}
\begin{tabular}{l c  ccc | ccc}
\toprule
\textbf{Marker} & \textbf{$\Delta_{\text{across}}$} &
\multicolumn{3}{c}{$\Delta_{\text{within}}^{\text{C-low}}$} &
\multicolumn{3}{c}{$\Delta_{\text{within}}^{\text{C-high}}$} \\
\cmidrule(lr){3-5} \cmidrule(lr){6-8}
& & Q2-Q1 & Q3-Q2 & Q4-Q3 & Q2-Q1 & Q3-Q2 & Q4-Q3 \\
\midrule
Anger       & \cellcolor{LightRed}39.8\textsuperscript{*}  
            & \cellcolor{LightGreen}–4.5 & \cellcolor{LightRed}49.1\textsuperscript{*} & \cellcolor{LightGreen}–18.2\textsuperscript{*}
            & \cellcolor{LightGreen}–7.6 & \cellcolor{LightRed}70.9\textsuperscript{*} & \cellcolor{LightGreen}–14.0\textsuperscript{*} \\
Griev.      & \cellcolor{LightRed}6.5\textsuperscript{*}  
            & \cellcolor{LightRed}2.4 & \cellcolor{LightRed}3.3 & \cellcolor{LightRed}0.01
            & \cellcolor{LightRed}0.6 & \cellcolor{LightRed}3.1 & \cellcolor{LightGreen}–0.8 \\
Power       & \cellcolor{LightRed}13.7\textsuperscript{*}  
            & \cellcolor{LightRed}4.5 & \cellcolor{LightGreen}–0.9 & \cellcolor{LightRed}2.9
            & \cellcolor{LightRed}7.6\textsuperscript{*} & \cellcolor{LightRed}3.4 & \cellcolor{LightRed}2.7 \\
Neg.        & \cellcolor{LightRed}7.5\textsuperscript{*}  
            & \cellcolor{LightRed}3.7\textsuperscript{*} & \cellcolor{LightRed}0.1 & \cellcolor{LightRed}1.0
            & \cellcolor{LightGreen}-1.4\textsuperscript{*} & \cellcolor{LightRed}4.1\textsuperscript{*} & \cellcolor{LightGreen}-0.1 \\
In-group    & \cellcolor{LightRed}9.5 
            & \cellcolor{LightRed}7.3 & \cellcolor{LightRed}2.8 & \cellcolor{LightGreen}–35.4
            & \cellcolor{LightGreen}–1.8 & \cellcolor{LightRed}5.2 & \cellcolor{LightGreen}–7.6 \\
Out-group   & \cellcolor{LightRed}22.0 
            & \cellcolor{LightGreen}–12.7 & \cellcolor{LightGreen}–42.8 & \cellcolor{LightGreen}–36.7
            & \cellcolor{LightRed}39.2 & \cellcolor{LightRed}5.1 & \cellcolor{LightGreen}–21.4 \\
\bottomrule
\end{tabular}
\caption{Percentage differences in production of content with markers of extremism. $\Delta_{\text{across}}$ shows the difference between \cext and \cneu channels over the year \update{(\cext - \cneu)}. $\Delta_{\text{within}}$ shows the marginal percentage differences within \cneu and \cext groups from one quarter to the next. Asterisks denote $p<0.05$.}
\label{tab:rq2}
\end{table}

\subsection{Results}\label{sec:rq2:results}
Similar to our analysis of consumption behaviors, we analyzed both baseline differences and temporal changes in linguistic markers for channels in \cext and \cneu. Our results are summarized in \Cref{tab:rq2}.

\para{Group-level differences.}
Overall, channels with high affinity to \ext participants produced content with significantly higher levels of multiple extremism-related markers compared to channels with affinity towards \neu participants. 
The most pronounced statistically significant differences were seen in anger (+39.8\%), power (+13.7\%), negative sentiment (+7.5\%), and grievance (+6.5\%).
These findings suggest that the media environment available to \ext users were not only shaped by user preferences but also by a unique production ecosystem that emphasized emotionally charged narratives more frequently than the ecosystem around \neu participants.

\para{Within-group differences.}
Although the temporal dynamics within groups are harder to interpret than the cross--group differences, several trends in Table \ref{tab:rq2} are notable. Most prominently, anger shows the largest increase for both groups during Q3 (49.1\% for \cneu and 70.9\% for \cext), followed by sharp declines in Q4 (-18.2\% and -14.0\%). We find that most markers rise in the middle of the year and fall back by the end.  
Anger, grievance, power, negative sentiment each show overall increase between Q4 and Q1, suggesting an upward trajectory in radical markers that points to a gradual intensification of radical tone within production.  
This channel composition helps explain the rise and drop, especially during Q3. Channels in \cext were predominantly news channels\footnote{We are referring to YouTube's \textit{News \& Politics} category.} (70\%) from across the political spectrum, including \textit{The Ben Shapiro Show}, \textit{MSNBC}, \textit{FOX}, and \textit{CNN}. In contrast, \cneu featured a relatively greater number of entertainment channels, such as \textit{Fail Army} and \textit{The Kelly Clarkson Show}, with fewer news channels (34\%). The language used in covering news events contributes to these fluctuations, reflecting the more reactive and emotionally charged narratives commonly found in political content.
Taken together, these within--group patterns suggest that while both groups respond to common temporal shocks, \cext channels are more reactive and more synchronized in their surges, especially in Q3. This clustering of anger, power, and grievance within \cext underscores a more coordinated and volatile production environment compared to the diffuse and uneven changes within \cneu.

\para{Takeaways.}
Although, \cext produced content that consistently exhibited higher levels of anger, power, negativity, and grievance markers, both \cext and \cneu showed similar increasing and decreasing trends throughout the year.
The fluctuations in production patterns are mirrored, and likely intertwined with, the consumption patterns observed in \Cref{sec:rq1}.
Taken together, these results support a production-side explanation for at least part of the differences in user consumption behaviors and suggest that content with higher rates of extremism-related markers were both consumed and produced differently.

\para{Contextualizing our findings.}
\update{We observe that production is episodic in nature with extreme highs and lows between quartiles (especially for anger in Q3). Whereas consumption is not as volatile. It shows a stable increase or decrease (within 20\%) for most markers.}
%
%
These findings gain additional significance when considered in relation to the sociopolitical context of the study period (July 2021-August 2022).
\update{\Cref{fig:topics} and \Cref{fig:users-anger} shows the evolution of anger scores (averaged weekly) across the year with summaries of the top topics derived from both the consumption and production trends.}
The early part of the study coincided with the delta wave of Covid-19 \cite{CDCMuseum_2024}, vaccine mandate controversies \cite{bardosh2022unintended}, multiple state-level bans on abortion \cite{Spitzer2021StateProgress, McCann2022AbortionTimes}, and the January 6th Capitol riot hearings \cite{GovInfo_2022}. 
%
%
In contrast, the second half of our study period which saw a universal increase of power and negative sentiment markers included coverage of the Russian invasion of Ukraine \cite{walker2023conflict} and the accompanying sharp increase in inflation and economic instability. 
A major difference in the production and consumption patterns was that fundamentally, the production channels' dominant themes were political and highly charged, with minor entries of entertainment—mostly satire. Whereas consumers, in addition to watching the political topics, showed a much better spread of entertainment. We saw topics such as music, cooking, religion etc. in all the quartiles resulting in less reactive peaks (as shown by the normalized anger scores) when compared to producers. 
%
%


\section{Production and consumption dynamics} \label{sec:rq3}

In this section, we investigate whether trends in content consumption influence trends in content production, or vice versa. Specifically, we ask {\em RQ3. What is the direction of the influence relationship between the content that is being produced and the content that is being consumed?} This question is motivated by the need to assess whether producers shape public discourse by introducing new content, or whether consumer demand drives the nature of content being created over time. We first describe the methods used for this analysis (\Cref{sec:rq3:methods}) and then present our findings.

\subsection{Methods} \label{sec:rq3:methods}
\update{We analyze four daily-aggregated time series: radical production (channels with \(\alpha(c, \mathcal{G}_{\text{ext}}) > 1.5\), see Eq.~\eqref{eq:alpha}), radical consumption (users in \(\mathcal{G}_{\text{ext}}\)), neutral production (channels with \(\alpha(c, \mathcal{G}_{\text{neu}}) > 1.5\)), and neutral consumption (users in \(\mathcal{G}_{\text{neu}}\))}. We aim to understand whether production behavior is shaped by prior consumption patterns, or whether content consumption simply follows what is being produced.
\update{To avoid artificially inflating the correlation between production and consumption, we excluded from the production time series any videos that were also present in the consumption time series.}

\para{Predicting influence.}
To assess the influence between content production and consumption, we employed Granger causality analysis \cite{granger}, a regression-based method that evaluates whether past values of one time series (e.g., production) improve the prediction of another (e.g., consumption), beyond what can be predicted from its own history. Specifically, we compared two linear regression models: a restricted model using only lagged values of the dependent series (e.g., consumption), and an unrestricted model that adds lagged values of the independent series (e.g., production). If the unrestricted model significantly reduces prediction error, as measured by the residual sum of squares, we compute an F-statistic to quantify this improvement. A significant F-score indicates that the additional predictors contribute meaningful information, suggesting that the independent time series Granger-causes the dependent one. This approach allows us to detect temporal precedence in the relationship, identifying whether producers are shaping consumer behavior or responding to it.

\para{Measuring directionality.}
We assess the directionality of influence between content consumption and content production by conducting two separate Granger causality tests: one testing whether past values of consumption can predict future production trends (\textit{consumption $\rightarrow$ production}), and the other testing the reverse (\textit{production $\rightarrow$ consumption}) for both \neu and \ext.
If only one of these directions yields a statistically significant result, it suggests a unidirectional influence—i.e., one process (either consumption or production) provides predictive information about the future trajectory of the other, while not being influenced in return. Conversely, if both directions show statistically significant predictive relationships, this implies the presence of a feedback loop between the two series. 
%

\subsection{Results} \label{sec:rq3:results}







\begin{table}[t]
\centering
\small
\setlength{\tabcolsep}{4pt}
\begin{tabular}{l cccc | cccc}
\toprule
\multirow{2}{*}{\textbf{Marker}} &
 \multicolumn{4}{c}{Extreme} &
 \multicolumn{4}{c}{Neutral} \\
\cmidrule(lr){2-5} \cmidrule(lr){6-9}
& P$\rightarrow$C & Lag & C$\rightarrow$P & Lag & P$\rightarrow$C & Lag & C$\rightarrow$P & Lag \\
\midrule
Anger       &
 3.20* & 3 &
 3.38* & 3 &
 1.32  & 4 &
 0.81  & 4 \\

Griev.   &
 1.63  & 1 &
 2.08  & 1 &
 1.78  & 4 &
 1.51  & 4 \\

Power       &
 2.77* & 5 &
 0.92  & 5 &
 1.62  & 3 &
 0.74  & 3 \\

Neg.   &
 0.49  & 3 &
 0.25  & 3 &
 0.96  & 5 &
 2.58* & 5 \\

In-group    &
 1.89  & 2 &
 1.78  & 3 &
 0.64  & 5 &
 0.10  & 5 \\

Out-group   &
 3.79* & 5 &
 1.02  & 4 &
 0.08  & 5 &
 3.22* & 5 \\
\bottomrule
\end{tabular}
\caption{The P$\rightarrow$C and C$\rightarrow$P columns show the F statistic for the SSR-based F-test in the Granger causality analysis, with the lag corresponding to the specification that yielded the lowest $p$-value among those reported. Asterisks denote $p<0.05$.}
\label{tab:rq3}
\end{table}







The results for granger causality analysis are shown in \cref{tab:rq3}. We discuss these results to evaluate which group (producers or consumers) is influencing the other more.

\para{Production $\rightarrow$  Consumption.} 
This analysis investigates whether linguistic properties of produced content can help predict the properties of the consumed content.
In the \ext group, we observe significant result for anger (lag=3, F=3.2), power (lag=5, F=2.8), and out-group (lag=5, F=3.8). The high predictive capability in the production trend is indicative of an influence that producers have over consumption patterns especially for the content where the aforementioned features are involved. It is also interesting to note that in all the \PC of \ext, anger has the shortest lag (3 days), implying that the consumers react to it faster than other markers. In comparison, the response to content highlighting power dynamics or out-group cues is slightly delayed (5 days). 
%
%
\update{Multiple markers show a significant \PC relationship within \ext group but not in \neu. This suggests that the types of production strategies that successfully drive audience engagement differ between the two groups, where the audience might be as responsive of the radical markers in the \ext but not as much in the \neu.}

\para{Consumption $\rightarrow$  Production.}
This analysis tests whether consumption patterns influence what gets produced, a notion aligned with feedback loops where producers respond to audience preferences or demand.
The only significant result for \CP analysis in \ext is for anger (lag=3, F=3.2). This implies that when users consume more anger-laden content, future production of such content increases, making consumers the driving force of such content. Notably, anger has the shortest lag across all comparisons in both groups, suggesting a faster and possibly more reactive production trend that closely tracks user interest. In contrast, the longer lag for \neu implies either slower adaptation by producers or weaker pressure from consumers to shape content in response to consumption. 
\update{For \neu, negative sentiment (lag=5, F=2.6) and out-group references (lag=5, F=3.2) show significant \CP relationships. These results suggest that when neutral audiences consume more negatively valenced or out-group–focused content, producers subsequently increase the supply of such material. The longer lag lengths, however, imply a slower feedback cycle relative to anger in \ext, pointing to a more gradual and possibly less intense process of adjustment. }
%
%
Interestingly, anger is the only marker in \ext where we observe a bidirectional relation between production and consumption. In such cases, producers appear to respond to the nature of user consumption, and consumers, in turn, adapt to the newly produced content. This mutual influence indicates a dynamic co-evolution, where production and consumption reinforce each other over time, potentially accelerating the spread or reinforcement of specific content types.

\para{Takeaways.} 
\update{Our findings reveal that content production remains the dominant force shaping consumption behavior in \ext, with stronger evidence for \PC effects. Specifically, anger, power, and out-group show significant links in the \PC direction, indicating that when producers emphasize these markers, audiences subsequently consume more of them. 
By contrast, in the \CP direction, \neu display more significant markers than \ext, with negative sentiment and out-group both exerting influence on production. This suggests that in neutral ecosystems, consumer demand is a primary driver of what gets produced, particularly for negative and boundary-defining content. }
Notably, anger is the only marker exhibiting a bidirectional relationship in \ext, consistent with a feedback loop in which angry content attracts further consumption, which in turn fuels more production. This reciprocal dynamic underscores anger’s central role in sustaining radical content flows, while other markers tend to operate more unidirectionally depending on the group context.

\section{Discussion } \label{sec:discussion}

\para{Consumption and production asymmetries are associated with ideological drift.}
Across our research questions, we observed asymmetries in the consumption, production, and reinforcement of content containing markers of extremism.
On the consumer side, we found that participants who became more ideologically extreme were more strongly immersed in content containing markers of anger and grievance. 
At the same time, even ideologically stable participants experienced similar consumption trends, albeit with lower magnitudes, over the course of a year rife with polarizing and sensitive sociopolitical issues at the forefront.
These asymmetries also extended to the production side. Here, channels with high affinity to our extreme group produced content (besides the content that was already consumed by our participants) with consistently higher levels of anger, power, and grievance.
This suggests that consumption choices within our groups are not just because of user preferences and selection biases, but also targeted production where producers are shaping and reinforcing information ecosystems.

\para{Audience capture is asymmetric.}
Our Granger temporal analysis shows that in most cases, changes in production trends among high affinity creators precede changes in consumption behaviors, suggesting that audiences are responding to already created information environments rather than shaping them.
However, for the anger marker, we find a bidirectional influence between production and consumption indicating a feedback loop between producers and consumers.
In the specific case of the anger marker, we find that users  appear to have a stronger influence on their creators than the inverse.
This dynamic aligns with audience capture where production is shaped to meet audience demands.
This dynamic is not observed with other markers or with the ideologically stable users in a meaningfully significant way.

\para{Markers of extremism in production and consumption are strongly associated with real-world events.}
On examination of the timeseries and content associated with each marker, we found increases to be strongly aligned with major events that occurred during our study.
Notably, peaks in anger and grievance in consumption behaviors and production patterns, for both groups of participants, occurred at the height of the Russian invasion of Ukraine and increasing inflation in the US.
These findings suggest that media ecosystems are responsive to the political climate within which they exist.

\para{Limitations}
Like all research, our work is not without limitations. 
First, despite consisting of over 1.1K users who were demographically representative of the US voting population, our study is subject to the standard limitations of panel subject participation: selective participation, attrition, and inconsistent tracking. 
To mitigate the effects of these limitations, we used strict filtering criteria and only included participants with consistently high engagement (based on daily activity recorded) throughout the study period in our data analysis.
However, it is still possible that our data are incomplete --- \eg if users only enabled tracking on one of their devices or turned off tracking during specific media consumption.
\update{Second, our analysis does not take into account participants’ engagement with other platforms or media modalities outside of YouTube. As such, we cannot fully account for the influence of cross-platform content exposure (e.g., Facebook, TikTok, news sites), which may also shape ideological attitudes.}
Further, our analysis was entirely reliant on text analysis and ignored the potential for visual markers that may be present in videos.
We utilize dictionary-based linguistic markers to ensure methodological transparency required for understanding the media ecosystem. Although transformer-based approaches might yield higher predictive accuracy, they do not offer the same interpretability needed to identify specific linguistic patterns in media discourse.
Our timeseries Granger analysis can only offer insights into temporal influence relationships and do not establish causality.
However, to ensure that our inferred influence relationships were robust, we took specific measures to prevent data pollution in our timeseries --- \eg we removed all videos from associated with our \cext and \cneu channels from users' watch histories prior to measuring influence.
\update{Finally, our analyses focused exclusively on participants who became more extreme over the study period. We did not examine users who became less extreme, though doing so would provide an important point of comparison and help clarify whether the dynamics we identify are unique to ideological escalation or reflect broader processes of attitudinal change. Future work should explicitly investigate these de-escalation trajectories.}

\section{Related Work} \label{sec:related-work}

Our analysis engages with and builds upon two intersecting research domains: online extremism and media effects. In this section, we describe prior work in these domains and place our findings in context.

\para{Online extremism and radicalization.}
Research to illuminate the process of radicalization through social media use has drawn significant attention over the past few years.
This prior work, from a number of disciplines, has provided an important foundation for understanding how extreme ideologies emerge online by showing how social media increases exposure to extreme content, facilitate entry into fringe communities, and normalize hatred and violence.
Much of the early research related to online radicalization focused on identifying radicals online in order to study their behaviors \cite{cohen2014detecting, grover2019detecting}.
Among the earliest to consider the role of platforms in radicalization, Tufekci \cite{tufekci2018youtube} proposed the idea that YouTube's recommendation system promoted increasingly extreme content to maximize user engagement. 
This idea gained further traction following an investigation by Roose \cite{roose2019making} which tracked one user's drift into the alt-right through YouTube content.
These anecdata were confirmed by Ribeiro \etal \cite{ribeiro2020auditing} who, through a large-scale audit of the YouTube recommendation algorithm, showed how users could find themselves on pathways to extremist content through algorithmic recommendations. 
Others have expanded this narrative by considering community-level and cross-platform dynamics \cite{habib2022making, habib2020act, cinelli2021echo}, and algorithmic amplification effects \cite{huszar2022algorithmic, chitra2020analyzing}.
\update{
Ledwich and Zaitsev (\citeyear{ledwich2019algorithmic}) found that YouTube’s algorithm tended to favor mainstream media content over radical or fringe content, suggesting a decline in “rabbit hole” effects. 
Hosseinmardi et al. (\citeyear{hosseinmardi2024causally}) go further to estimate the causal impact of YouTube’s recommender system using a novel “counterfactual bots” method, which mimic real-user behavior to explore recommendations. Their work challenges earlier claims of algorithmic radicalization, instead highlighting user preference as the dominant force in shaping content exposure post-2019.
Similarly in another experiment, Hosseinmardi et al. (\citeyear{hosseinmardi2021examining}) used large-scale browsing data and showed that most viewers of extreme content found such videos via direct navigation or external links, rather than recommendations.
Chen et al. (\citeyear{chen2023subscriptions}) build on these findings by pairing behavioral data with survey responses and showing that exposure to alternative and extremist content is concentrated among users with pre-existing high levels of gender or racial resentment. They emphasize the role of
subscriptions and external referrers (as opposed to the algorithm) as the dominant pathways to extreme content.
Ribeiro et al. (\citeyear{ribeiro2023amplification}) introduce the concept of the “amplification paradox”, where algorithmic audits suggest that recommender systems promote extreme content, but real-world data shows that users rarely consume it unless it aligns with their preferences. This challenges simplistic notions of algorithmic radicalization by emphasizing the central role of user agency and demand, aligning with a broader supply-and-demand view of online extremism (\citeyear{munger2022right}).
A complementary perspective is offered by Lewis (\citeyear{lewis2018alternative}) in The \textit{Alternative Influence Network}, which maps a web of ideological influencers who collaborate and cross-promote within the YouTube ecosystem, normalizing radical viewpoints through parasocial relationships and reactionary commentary.
}
Our research extends and differs from this literature in two important ways.
First, by considering the behavioral data of participants that do exhibit increased tendencies towards more extreme ideologies, our findings avoid the representational limitations of sock puppets or online pseudonymous identities.
Further, we examine extremism as a process that is influenced by the consumption behaviors (demand) and production patterns (supply) that are incentivized by platforms.
In doing so, we are able to show that extremism is not just algorithmically delivered but shaped by an asymmetric information ecosystem.

\para{Media effects.}
Media effects research has long examined how content shapes public opinion, individual behaviors, and ideologies.
Early work on the subject argued that media informs and structures what people think and how they think about it \cite{scheufele2007framing, mccombs2002news}.
This concept has been empirically demonstrated in several more recent studies. 
By combining digital histories with self-reports, Guess \etal \cite{guess2018selective} demonstrated a relationship between political media consumption and belief systems.
Bail \etal \cite{bail2018exposure} extended this work through a controlled experiment which showed that the dynamics between media diets and ideology formation were not straightforward. They experimentally evaluated the effects of media diet on political polarization and demonstrated that cross-partisan media diets can lead to increased polarization. 
Other work has emphasized that cognitive biases, temporal effects, and emotional valence of media all play an important role in the creation of discourse and ideology \cite{kubin2021role, jenkins2016youth}.
We contribute to this body of work by empirically modeling content production and consumption as a co-evolving process.
Our Granger analysis shows that for most markers of extremism, production precedes consumption. However, in the case of `anger', we find that the direction is inverted particularly with users who later adopted more extreme ideologies.
This is aligned with the concept of `audience capture' \cite{lewis2020news} where creators take a reactionary approach by reflecting their audiences' opinions and ideologies. Thus, they are shaped by their audience rather than shaping their audience --- an inversion of the dynamics of the traditional media ecosystem.

\section{Conclusions}
Our analysis highlights some of the relationship between platform users, content creators, and characteristics of content.
We found that the group of participants who demonstrated more extreme changes in their ideologies were more consistently consuming content with significantly higher markers of anger and grievance.
This is likely a consequence of the mechanisms of platforms which reward creators for audience attention and engagement.
Creators responding to these signals creates a feedback loop where content is created to elicit engagement through emotional intensity and their consumers' responsiveness, which occurs due to their cognitive and selection biases, to this communication strategy is rewarded by the platform.
Our work suggests that current moderation approaches which focus on simply removing `harmful content' overlooks these dynamics due to poorly formed incentive structures.
Our work emphasizes the importance for platforms and researchers to assess the role of platform incentives, algorithmic feedback, system design, and cognitive biases on the information environment.

\bibliography{aaai25}
\clearpage
\subsection{Ethics Checklist}

\begin{enumerate}

\item For most authors...
\begin{enumerate}
    \item  Would answering this research question advance science without violating social contracts, such as violating privacy norms, perpetuating unfair profiling, exacerbating the socio-economic divide, or implying disrespect to societies or cultures?
    \answerYes{Yes, this study in increasing transparency to understand the effects of social media and will hopefully be a benefit to the society.}
  \item Do your main claims in the abstract and introduction accurately reflect the paper's contributions and scope?
    \answerYes{Yes.}
   \item Do you clarify how the proposed methodological approach is appropriate for the claims made? 
    \answerYes{Yes, each research question asked has its own method and justification provided.}
   \item Do you clarify what are possible artifacts in the data used, given population-specific distributions?
    \answerYes{Yes, we mention it in the section where we describe our dataset}
  \item Did you describe the limitations of your work?
    \answerYes{Yes, see \Cref{sec:discussion}}
  \item Did you discuss any potential negative societal impacts of your work?
    \answerNo{No, because our study is aimed understanding how social media impacts our lives. It is not an experiment/product impacting people, rather bringing transparency by understand the phenomena that's already happening around us.}
      \item Did you discuss any potential misuse of your work?
    \answerNo{No, currently we are unaware of any potential misuse, but we will promptly inform authorities, if we come across any.}
    \item Did you describe steps taken to prevent or mitigate potential negative outcomes of the research, such as data and model documentation, data anonymization, responsible release, access control, and the reproducibility of findings?
    \answerYes{Yes, our research used panel data of the U.S population, which is why we are only releasing the results to ensure participant anonymity.}
  \item Have you read the ethics review guidelines and ensured that your paper conforms to them?
    \answerYes{Yes}
\end{enumerate}

\item Additionally, if your study involves hypotheses testing...
\begin{enumerate}
  \item Did you clearly state the assumptions underlying all theoretical results?
    \answerNA{NA}
  \item Have you provided justifications for all theoretical results?
    \answerNA{NA}
  \item Did you discuss competing hypotheses or theories that might challenge or complement your theoretical results?
    \answerNA{NA}
  \item Have you considered alternative mechanisms or explanations that might account for the same outcomes observed in your study?
    \answerNA{NA}
  \item Did you address potential biases or limitations in your theoretical framework?
    \answerNA{NA}
  \item Have you related your theoretical results to the existing literature in social science?
    \answerNA{NA}
  \item Did you discuss the implications of your theoretical results for policy, practice, or further research in the social science domain?
    \answerNA{NA}
\end{enumerate}

\item Additionally, if you are using existing assets (e.g., code, data, models) or curating/releasing new assets, \textbf{without compromising anonymity}...
\begin{enumerate}
  \item If your work uses existing assets, did you cite the creators?
    \answerNA{NA}
  \item Did you mention the license of the assets?
    \answerNA{NA}
  \item Did you include any new assets in the supplemental material or as a URL?
    \answerNA{NA}
  \item Did you discuss whether and how consent was obtained from people whose data you're using/curating?
    \answerYes{Yes, we are using YouGov panel data. Consent was obtained from all participants. They also had the option to stop participating at all times without any penalty. It is discussed in the paper.}
  \item Did you discuss whether the data you are using/curating contains personally identifiable information or offensive content?
    \answerYes{Yes, we discuss the data anonymity in \Cref{sub:ethics}}
\item If you are curating or releasing new datasets, did you discuss how you intend to make your datasets FAIR (see \citet{fair})?
\answerNA{NA}
\item If you are curating or releasing new datasets, did you create a Datasheet for the Dataset (see \citet{gebru2021datasheets})? 
\answerNA{NA}
\end{enumerate}

\item Additionally, if you used crowdsourcing or conducted research with human subjects, \textbf{without compromising anonymity}...
\begin{enumerate}
  \item Did you include the full text of instructions given to participants and screenshots?
    \answerYes{No, but we do mention the scales used in our survey in the dataset section}
  \item Did you describe any potential participant risks, with mentions of Institutional Review Board (IRB) approvals?
    \answerYes{Yes, they were made aware of the potential risks of participation, and had the option to stop participation at all times.}
  \item Did you include the estimated hourly wage paid to participants and the total amount spent on participant compensation?
    \answerNA{Yes}
   \item Did you discuss how data is stored, shared, and deidentified?
   \answerNo{No, because we received anonymized data from YouGov. We stored the data in our firewall protected server, which is only accessible to the authors of the paper.}
\end{enumerate}

\end{enumerate}
\begin{figure*}[t!]
    \centering
    \begin{subfigure}[t]{0.48\textwidth}
        \centering
        \includegraphics[width=\linewidth]{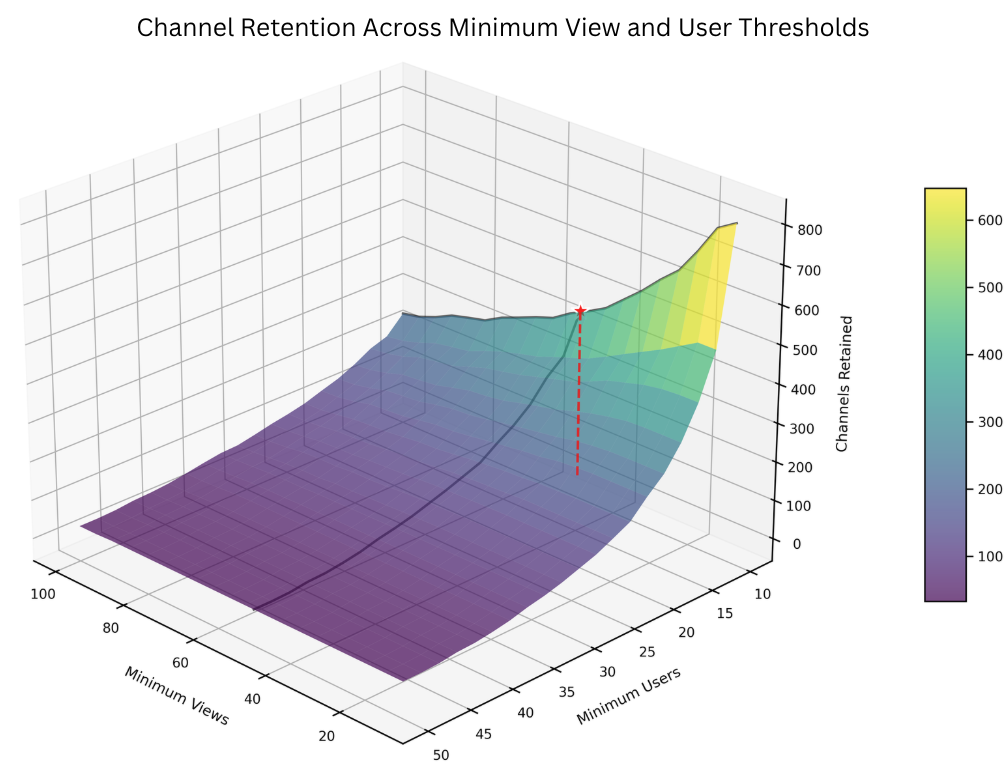}
        \caption{Channel retention vs. views and users.}
        \label{fig:channel-association}
    \end{subfigure}
    \hfill
    \begin{subfigure}[t]{0.48\textwidth}
        \centering
        \includegraphics[width=\linewidth]{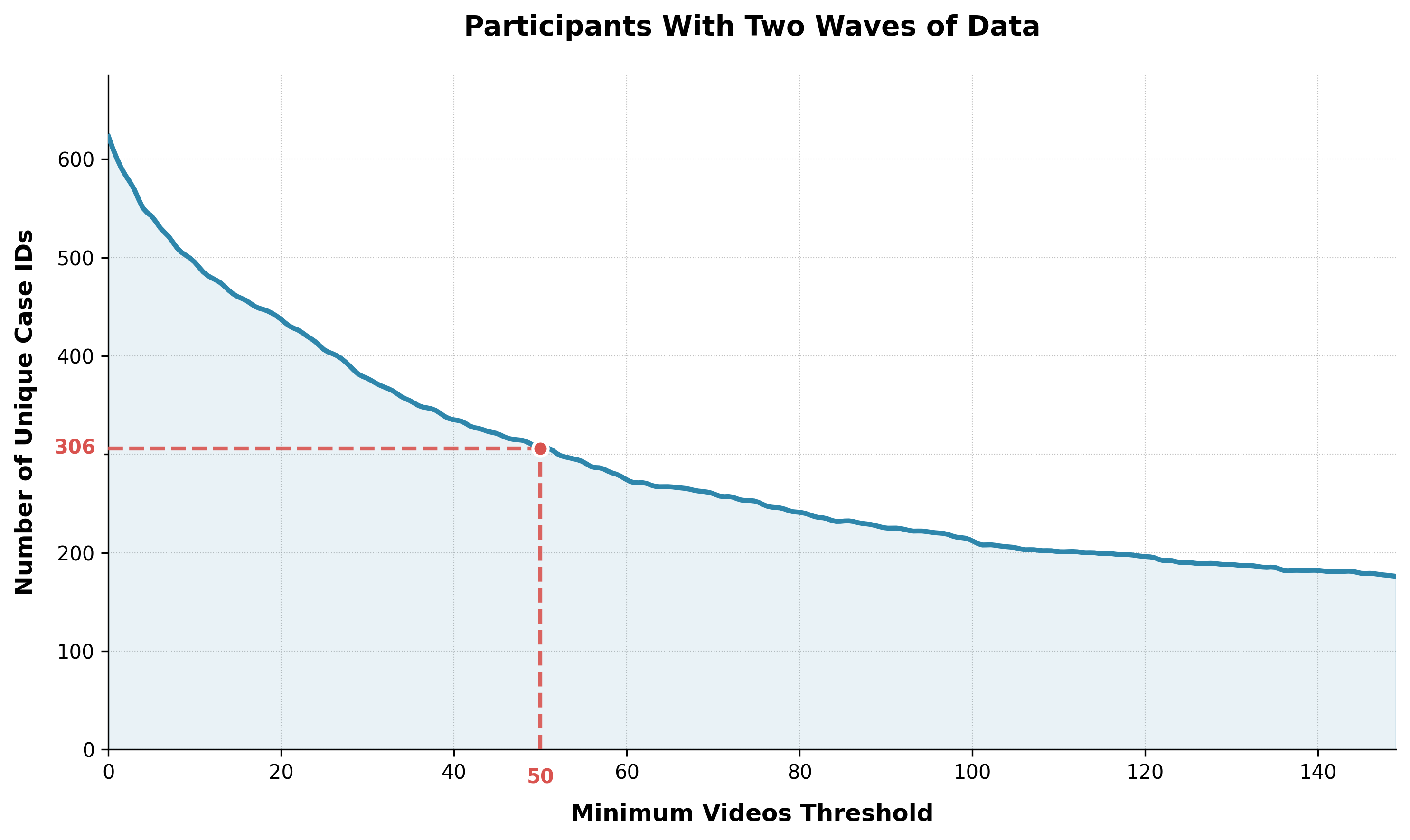}
        \caption{Participant eligibility vs. video threshold.}
        \label{fig:threshold-two}
    \end{subfigure}
    
    \caption{Sensitivity analysis for data inclusion. Plot (a) illustrates the tradeoff between engagement filters and channel sample size, while (b) shows the impact of video count thresholds on participant retention (selected threshold $x=50$ marked in red).}
    \label{fig:combined-thresholds}
\end{figure*}

\newpage

\section{Appendix} \label{sec:appendix}


\subsection{Thresholds} \label{appdendix:threhsold}
\paragraph{Selecting participants.}
We began with 1,896 panel participants, of which 1,100 completed both survey waves. From these, 623 had at least watched one YouTube video over the one-year observation window. To ensure a meaningful level of behavioral data per user, we varied the threshold for minimum videos watched and plotted the number of participants retained at each level (see \Cref{fig:threhsold-two}). As the curve shows, there is no obvious inflection point or ``elbow''—the trade-off between sample size and data richness is continuous. We ultimately selected a cutoff of 50 videos, which preserves roughly 250 participants. This threshold corresponds to about one video per week, which we consider a minimal but reasonable signal of sustained platform use. The threshold is slightly above the sample median of 47 videos.

\paragraph{Selecting channels.}
For the production-side analysis, our goal was to include only those channels with sufficient engagement to support meaningful comparisons between groups. Specifically, we filtered out channels that had low total views or were watched by very few unique users. \Cref{fig:channel-association} shows how many channels meet various combinations of minimum view and minimum user thresholds. The top-right corner (in yellow) represents channels that were widely and frequently viewed, while the bottom-left corresponds to fringe or sparsely viewed channels. Based on a sweep through this space and manual inspection of affinity results, we selected a threshold of at least 10 unique users and 50 total views per channel. This threshold excludes channels whose inclusion would inflate group-level metrics based on a single user’s binge consumption, while retaining those with meaningful group-level signal.


\begin{figure*}[t!]
    \centering
    \includegraphics[width=1\linewidth]{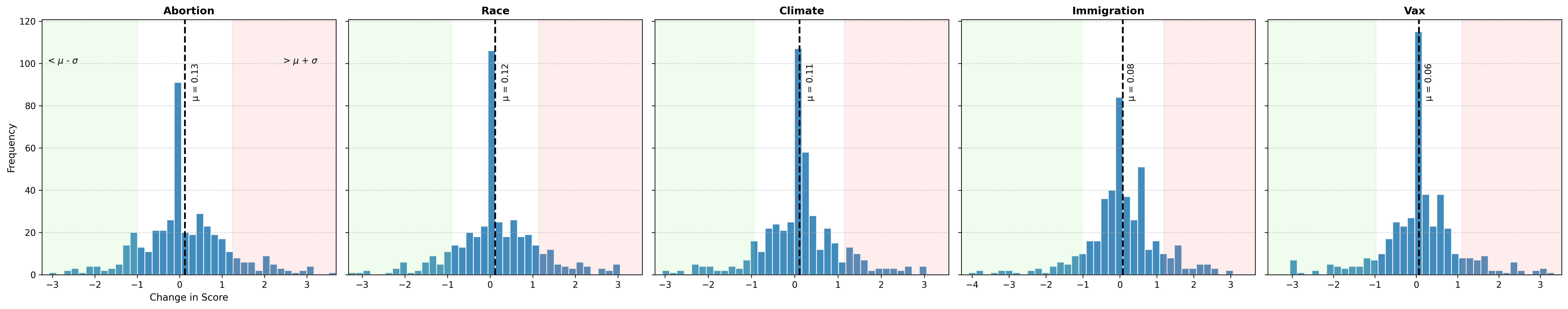}
    \caption{Issue specific shifts in attitude scores over the year.}
    \label{fig:issue-specific}
\end{figure*}

\subsection{Issue Specific Changes in Ideologies} \label{appendix:specific-issues}

\Cref{fig:issue-specific} shows that the ideological shift of the survey participants broken down by issues. Abortion had the highest mean shift (0.13) while Vaccination had the lowest (0.06). However, an important trend that all the issues share is that all of them observed a positive shift, even if the magnitude is small. Further, we see that the distributions are tightly centered around zero, consistent with prior findings that extremist attitudes are generally stable over year-long intervals \cite{nivette2022understanding, ansolabehere2008strength}.

\subsection{Topic Modeling} \label{appendix:topic-model}

\paragraph{Setup.}
We analyzed production and consumption corpora in four temporal quartiles (Q1--Q4). For each quartile and corpus, we fitted \textsc{BERTopic} \cite{grootendorst2022bertopic}. To find the optimal number of topics, we performed a grid search over \textit{min\_cluster\_size} $\in \{50,60,70,80,90,100\}$ (step size 10). This parameter enforces a minimum size for any discovered cluster. We measured the topic quality based on two metrics:

\begin{enumerate}
     
    \item\textbf{Topic Coherence} ($C$): We evaluate topic quality using gensim's \textit{CoherenceModel}\footnote{\url{https://radimrehurek.com/gensim/models/coherencemodel.html}}. This metric measures the semantic consistency of the top words in each topic by assessing their pairwise co-occurrence (using Normalized Pointwise Mutual Information) in the corpus. Higher coherence values indicate that the top words of a topic frequently appear together in similar contexts, making the topic more coherent, while lower values suggest incoherent groupings.

    \item\textbf{Silhouette} ($S$): The silhouette score quantifies clustering quality by comparing how close each data point is to others in its assigned cluster relative to points in the nearest neighboring cluster. For a document $i$, it is defined as:
    \[
    s(i) = \frac{b(i) - a(i)}{\max \{a(i), b(i)\}}
    \]
    where $a(i)$ is the average distance to all points in the same cluster, and $b(i)$ is the lowest average distance to points in another cluster. The score ranges from $-1$ to $+1$, with higher values indicating that points are well matched to their own cluster and clearly distinct from others, while values near $0$ or negative suggest overlapping or misassigned clusters.
\end{enumerate}

We select the min\_cluster\_size that maximizes the harmonic mean of the two normalized metrics:
\[
\mathrm{HM}_k \;=\; \frac{2}{\frac{1}{\hat{C}_k}+\frac{1}{\hat{S}_k}}.
\]

We associate the highest value of harmonic mean with the optimal value of min\_cluster\_size per quartile for each group. The themes in Fig.~\ref{fig:topics} are defined by keywords and the titles associated with the top topics.

\end{document}